# Types of Cost in Inductive Concept Learning

**Peter Turney**  PETER.TURNEY@NRC.CA
Institute for Information Technology, National Research Council of Canada, M-50 Montreal Road, Ottawa, Ontario, Canada, K1A 0R6

## Abstract

Inductive concept learning is the task of learning to assign cases to a discrete set of classes. In real-world applications of concept learning, there are many different types of cost involved. The majority of the machine learning literature ignores all types of cost (unless accuracy is interpreted as a type of cost measure). A few papers have investigated the cost of misclassification errors. Very few papers have examined the many other types of cost. In this paper, we attempt to create a taxonomy of the different types of cost that are involved in inductive concept learning. This taxonomy may help to organize the literature on cost-sensitive learning. We hope that it will inspire researchers to investigate all types of cost in inductive concept learning in more depth.

## 1. Introduction

This paper is an attempt to list the different costs that may be involved in inductive concept learning. The paper assumes the standard inductive concept learning scenario. We have a set of cases (i.e., examples, vectors, observations) represented as vectors in an abstract space of features (i.e., tests, measurements, sensor values, attribute values). Each case belongs to a class (i.e., the feature space is partitioned into a finite set of distinct subsets; there is a function mapping from feature space into a finite set of symbols). The learning algorithm generates hypotheses that may be used to predict the class of new cases.

In the following, "cost" should be interpreted in its most abstract sense. Cost may be measured in many different units, such as monetary units (dollars), temporal units (seconds), or abstract units of utility (utils). In medical diagnosis, cost may include such things as the quality of life of the patient, in so far as such things can be (approximately) measured. In image recognition, cost might be measured in terms of the CPU time required for certain computations. We take "benefit" to be equivalent to negative cost.

Often we are uncertain about costs. We can represent this uncertainty with a probability distribution over a range of possible costs. This applies to all of the following costs. In this paper, for ease of exposition, we will assume that we are certain about costs.

## 2. Cost of Misclassification Errors

Suppose there are $C$ classes. In general, we may have a $C$ x $C$ matrix, where the element in row $i$ and column $j$ specifies the cost of assigning a case to class $i$, when it actually belongs in class $j$. Typically (but not necessarily) the cost is zero when $i$ equals $j$. In a minor variation on this approach, we may have a rectangular matrix, where there is an extra row for the cost of assigning a case to the *unknown* (or "too-difficult-for-this-learner") class.

### 2.1 Constant Error Cost

The cost of a certain type of error (the value of a cell in the cost matrix) may be a constant (the same value for all cases). This is the most commonly investigated type of cost; for example, see Breiman *et al.* (1984) or Hermans *et al.* (1974).

If the cost is zero if $i$ equals $j$ and one otherwise, then our cost measure is the familiar *error-rate* measure. If the cost is one if $i$ equals $j$ and zero otherwise, then our cost measure (in this case, our "benefit measure") is the familiar *accuracy* measure.

### 2.2 Conditional Error Cost

The cost of a certain type of error may be conditional on the circumstances.

#### 2.2.1 Error Cost Conditional on Individual Case

The cost of a classification error may depend on the nature of the particular case. For example, in detection of fraud, the cost of missing a particular case of fraud will depend on the amount of money involved in that particular case (Fawcett and Provost, 1996, 1997). Similarly, the cost of a certain kind of mistaken medical diagnosis may be conditional on the particular patient who is misdiagnosed. For example, the misdiagnosis may be more costly in elderly patients.

It may be possible to represent this situation with a constant error cost by distinguishing sub-classes. For example, instead of two classes, "sick" and "healthy", there could be three classes, "sick-and-young", "sick-and-elderly", and "healthy". This is an imperfect solution when the cost varies continuously, rather than discretely.

2.2.2 ERROR COST CONDITIONAL ON TIME OF CLASSIFICATION

In a time-series application, the cost of a classification error may depend on the timing. Consider a classifier that monitors sensors that measure a complex system, such as a manufacturing process or a medical device. Suppose that the classifier is intended to signal an alarm if a problem has occurred or will soon occur. The sensor readings must be classified as either "alarm" or "no-alarm". The cost of the classification depends on whether the classification is correct and also on the timeliness of the classification. The alarm is not useful unless there is sufficient time for an adequate response to the alarm (Fawcett and Provost, 1996, 1997, 1999).

Again, it may be possible to represent this situation with a constant error cost by distinguishing sub-classes. Instead of two classes, "alarm" and "no-alarm", there could be "alarm-with-lots-of-time", "alarm-with-a-little-time", "alarm-with-no-time", and "no-alarm". Again, this is an imperfect solution when the cost varies continuously as a function of the timeliness of the alarm.

2.2.3 ERROR COST CONDITIONAL ON CLASSIFICATION OF OTHER CASES

In some applications, the cost of making a classification error with one case may depend on whether errors have been made with other cases. The familiar *precision* and *recall* measures, widely used in the information retrieval literature, may be seen as cost measures of this type (van Rijsbergen, 1979). For example, consider an information retrieval task, where we are searching for a document on a certain topic. Suppose that we would be happy if we could find even one document on this topic. If we are given a collection of documents to classify as "relevant" or "not-relevant" for the given topic, then the cost of mistakenly assigning a relevant document to the not-relevant class depends on whether there are any other relevant documents that we have correctly classified.

As another example, in activity monitoring, if you issue an alarm twice in succession for the same problem, the benefit of the second alarm is less than the benefit of the first alarm, assuming both alarms are correct classifications (Fawcett and Provost, 1999). This is related to Section 2.2.2.

2.2.4 ERROR COST CONDITIONAL ON FEATURE VALUE

The cost of making a classification error with a particular case may depend on the value of one or more features of the case.

## 3. Cost of Tests

Each test (i.e., attribute, measurement, feature) may have an associated cost. For example, in medical diagnosis, a blood test has a cost.

Turney (1995a) points out that we can only rationally determine whether it is worthwhile to pay the cost of a test when we know the cost of misclassification errors. If the cost of misclassification errors is much greater than the cost of tests, then it is rational to purchase all tests that seem to have some predictive value. If the cost of misclassification errors is much less than the cost of tests, then it is not rational to purchase any tests.

### 3.1 Constant Test Cost

The cost of performing a certain test may be a constant. Each test has a different cost, but the cost of a given test is the same for all cases (Núñez, 1988, 1991; Tan, 1991a, 1991b, 1993).

### 3.2 Conditional Test Cost

The cost of performing a certain test may be conditional on the circumstances surrounding the test.

3.2.1 TEST COST CONDITIONAL ON PRIOR TEST SELECTION

The cost of performing a certain test on a given patient may be conditional on the previous tests that have been chosen for the patient. For example, a group of blood tests ordered together may be cheaper than the sum of the costs of each test considered by itself, since the tests share common costs, such as the cost of collecting blood from the patient (Turney, 1995a).

3.2.2 TEST COST CONDITIONAL ON PRIOR TEST RESULTS

The cost of performing a certain test on a patient may be conditional on the results of previous tests. For example, the cost of a blood test is conditional on the patient's age. Thus a blood test must be preceded by a "patient-age" test, which determines the cost of the blood test.

3.2.3 TEST COST CONDITIONAL ON TRUE CLASS OF CASE

The cost of performing a certain test on a patient may be conditional on the correct diagnosis of the patient. For example, the cost of an exercise stress test on a patient may be conditional on whether the patient has heart disease. The stress test could cause heart failure, which adds to the total cost of the test.

3.2.4 TEST COST CONDITIONAL ON TEST SIDE-EFFECTS

The cost of performing a certain test on a patient may be conditional on possible side-effects of the test. For example, some patients are allergic to the dies that are used in certain radiological procedures. One side-effect of such a radiological test is an allergic reaction, which may increase the cost of the test.

3.2.5  TEST COST CONDITIONAL ON INDIVIDUAL CASE

The cost of performing a certain test may depend on idiosyncratic properties of the individual case.

3.2.6  TEST COST CONDITIONAL ON TIME OF TEST

The cost of performing a certain test may depend on the timing of the test.

## 4. Cost of Teacher

Suppose we have a practically unlimited supply of unclassified examples (i.e., cases, feature vectors), but it is expensive to determine the correct class of an example. For example, every human is a potential case for medical diagnosis, but we require a physician to determine the correct diagnosis for each person. A learning algorithm could seek to reduce the cost of teaching by actively selecting cases for the teacher. A wise learner would classify the easy cases by itself and reserve the difficult cases for its teacher.

If a learner has no choice in the cases that it must classify, then it can only rationally determine whether it should pay the cost of a teacher when it knows the cost of misclassification errors. A rational learner would, for each new case, calculate the expected cost of classifying the case by itself versus the cost of asking a teacher to classify the case. This scenario can be handled by using a rectangular cost matrix, as we discussed in Section 2.

In a more interesting scenario, the learner can explore a (possibly infinite) set of unclassified (unlabelled) examples and select examples to ask the teacher to classify. This kind of learning problem is known as *active learning*. In this scenario, we can rationally seek to minimize the cost of the teacher even when we do not know the cost of misclassification errors, if we assume that asking the teacher costs more than a correct classification (otherwise you would always ask the teacher) but less than an incorrect classification (otherwise you would never ask the teacher). However, we may be able to make better decisions if we have more information about the cost of misclassification errors.

### 4.1 Constant Teacher Cost

In the simplest situation, the cost of asking a teacher to classify a case is assumed to be the same for all cases. This is the usual assumption in the active learning literature (Cohn *et al.*, 1995, 1996; Krogh and Vedelsby, 1995; Hasenjager and Ritter, 1998).

### 4.2 Conditional Teacher Cost

In a more complex situation, the cost of asking a teacher to classify a case may vary with the circumstances of the case. For example, the cost may increase with the complexity of the case. On the other hand, the teacher may choose to penalize the student for asking the class of a trivial case.

## 5. Cost of Intervention

Suppose we have data from a manufacturing process. Each feature might be a measurement of an aspect of the process, while the classes might be different types of products. A learning algorithm could induce rules that predict the type of product, given the corresponding features. Suppose we wish to intervene in the manufacturing process, to make more of one type of product. We could give the induced rules a causal interpretation.

For example, assume that we have a continuous process, such as petroleum distillation. Suppose a rule says, "If sensor A has a value greater than B, then the yield of product type C will increase." If this rule has causal significance, then we may be able to increase the amount of product type C by intervening in the process so that sensor A consistently has a value greater than B. There may be a cost associated with this intervention. Each feature may have a corresponding cost, where the cost represents the effort required to intervene in the manufacturing process at the particular point represented by the feature (Verdenius, 1991).

This is somewhat different from the idea of assigning a cost to a feature based on the effort required to measure the feature. Instead, the cost represents the effort required to manipulate the process in order to alter the feature's value.

### 5.1 Constant Intervention Cost

In the simplest scenario, the cost of intervention for a given feature is the same for all cases (Verdenius, 1991).

### 5.2 Conditional Intervention Cost

In a more complex scenario, the cost of intervention for a feature may depend on the particular case (for a continuous process, "observation" may be a more appropriate term than "case"). For example, if a sensor is observed to be near its average value, it may be relatively easy to manipulate the process in order to move the average up or down slightly. However, if the sensor is observed to be far from its average value, it may be quite difficult to move it even further from its average value (van Someren *et al.,* 1997).

## 6. Cost of Unwanted Achievements

When we are dealing with the scenario described in Section 5, where induced rules are used to intervene in a causal process, the nature of misclassification error cost changes. Suppose a rule says, "If sensor A has a value greater than B, then the yield of product type C will increase." If we are using this rule to make predictions, then there is a misclassification error cost associated with incorrect predictions. If we are using this rule to intervene in a manufacturing process, then there is a similar cost associated with "unwanted achievements" (van Someren *et al.,* 1997).

Suppose that the rule makes successful predictions for 90% of the cases for which the antecedent of the condition ("sensor A has a value greater than B") is satisfied. If we can give the rule a causal interpretation, then we may expect that, if we manipulate the process so that sensor A is always greater than B, then the yield of product type C will increase 90% of the time. The remaining 10%, where our manipulation fails to increase the yield of product type C, are a cost of using this rule. These 10% are "unwanted achievements" of the rule (van Someren *et al.,* 1997). (The terminology "unwanted achievements" seems somewhat odd, but this is the terminology used in van Someren *et al.,* 1997, and we are reluctant to confuse the issue by introducing new terminology.)

### 6.1 Constant Unwanted Achievement Cost

If the cost of unwanted achievements is constant, then we can use a cost matrix, as with the cost of misclassification errors (Section 2).

### 6.2 Conditional Unwanted Achievement Cost

The cost of unwanted achievements may vary with factors such as the market demand for the unwanted achievement, the cost of disposing of the unwanted achievement, the cost of repairing or refining the unwanted achievement, or the quantity of the unwanted achievement.

## 7. Cost of Computation

Computers are a limited resource, so it is meaningful to consider the cost of computation. The various types of computational complexity are essentially different forms of cost that we may wish to take into account.

We may distinguish the cost of computation by whether it is static or dynamic, or by whether it is incurred during training or during testing.

### 7.1 Static Complexity

A computer program, considered as a static object, has a measurable complexity.

#### 7.1.1 SIZE COMPLEXITY

The size complexity of a computer program may be measured in several ways, such as the number of lines of code, or the number of bytes. Since the code takes up memory space in the computer, there is clearly a corresponding cost.

Turney (1995b) shows how it is possible, under certain circumstances, to treat size complexity as a kind of test cost (Section 3). In this case, each feature that is to be measured corresponds to a block of code that computes the feature. The cost of measuring a feature is proportional to the size of the corresponding block of code. The goal is to minimize the total size of the code, which is approximately the same as minimizing the total cost of the features. (It is only approximate, because blocks of code can combine in non-additive ways.)

#### 7.1.2 STRUCTURAL COMPLEXITY

The structural complexity of a computer program might be measured by the number of loops in the program, the depth of nesting of the loops, or the number of recursive function calls. Structural complexity has a cost; for example, software with high structural complexity is more difficult for software engineers to maintain.

### 7.2 Dynamic Complexity

The dynamic complexity of a program is the execution time or memory space consumed by the program. Unlike static complexity, dynamic complexity is a function of the input to the program.

#### 7.2.1 TIME COMPLEXITY

Time complexity may be measured in many different ways. Even with a specific architecture, there are many possible choices. For example, the time complexity of a Turing machine might be measured by the number of movements of the read/write head, by the number of direction changes of the read/write head, or by the number of state transitions of the finite state machine.

For example, a learning algorithm that discovers new features may take into account the time complexity of calculating the new features (Fawcett, 1993). In this example, time complexity is a kind of test cost (Section 3).

#### 7.2.2 SPACE COMPLEXITY

The space complexity of a program is the amount of memory it requires for execution with a given input. Clearly memory has a cost. There are well-known (in the theory of computational complexity) trade-offs between time complexity and space complexity.

### 7.3 Training Complexity

The cost of computational complexity may be incurred during training, when the algorithm is learning to classify.

### 7.4 Testing Complexity

The cost of computational complexity may be incurred during testing, when the algorithm is making predictions.

Case-based reasoning, for example, typically has a low dynamic complexity during training, but a high dynamic complexity during testing. On the other hand, neural networks typically have a high dynamic complexity during training, but a low dynamic complexity during testing.

## 8. Cost of Cases

There is often a cost associated with acquiring cases (i.e., examples, feature vectors). Typically a machine learning researcher is given a small set of cases, and acquiring further cases is either very expensive or practically impossible. This is why many papers are concerned with the "learning curve" (performance as a function of the sample size) of a machine learning algorithm.

### 8.1 Cost of Cases for a Batch Learner

Suppose that we plan to use a batch learning algorithm to build a model that will be embedded in a certain software system. The model will be built once, using a set of training data. The software system will perform some task, using the embedded model, a certain number of times over the operational lifetime of the system.

For a given learning algorithm, if we can estimate (1) the learning curve (the relation between training set size and misclassification error rate), (2) the expected number of classifications that the learned model will make when embedded in the operational system, over the lifetime of the system, (3) the cost of misclassification errors, and (4) the cost of acquiring cases for training data, then we can calculate the combined cost of training (building the model) and operating (using the model) as a function of training set size. We can then optimize the size of the training set to minimize this combined cost (Provost *et al.*, 1999).

Alternatively, an adaptive learning system, given (1) the expected number of classifications that the learned model will make when embedded in the operational system, (2) the cost of misclassification errors, and (3) the cost of acquiring cases for training data, could adjust its learning curve (fast but naïve versus slow but sophisticated) and training set size to optimize the combined cost of training and operating.

### 8.2 Cost of Cases for an Incremental Learner

Suppose that we plan to use an incremental learning algorithm to build a model that will be embedded in a certain software system. Unlike the batch learning scenario, the model will be continuously refined over the operational lifetime of the system. However, it is likely that the software system cannot be operationally deployed without any training. We must decide how many training cases we should give to the incremental learner before it becomes sufficiently reliable to deploy the software system. To make this decision rationally, we need to assign a cost to acquiring cases for training data. The situation is similar to the batch learning situation, except that we suppose that the misclassification error rate will continue to decrease after the software system is deployed.

## 9. Human-Computer Interaction Cost

There is a human cost to using inductive learning software. This cost includes finding the right features for describing the cases, finding the right parameters for optimizing the performance of the learning algorithm, converting the data to the format required by the learning algorithm, analyzing the output of the learning algorithm, and incorporating domain knowledge into the learning algorithm or the learned model.

### 9.1 HCI Cost of Data Engineering

By "data engineering", we mean the steps required to prepare the data so that they are suitable for a standard inductive concept learning algorithm. This includes finding the right features and converting the data to the required format. Although there has been some discussion of the issues involved in data engineering (Turney *et al.*, 1995), we are not aware of any attempt to measure the HCI costs involved in data engineering.

### 9.2 HCI Cost of Parameter Setting

Most learning algorithms have a number of parameters that effect their performance, often by adjusting their bias. There is a cost involved in determining the best parameter settings. Often cross-validation is used to set the parameters (Breiman *et al.*, 1984). Again, we are not aware of any attempt to measure the HCI costs of parameter setting.

### 9.3 HCI Cost of Analysis of Learned Models

There is a human cost associated with understanding induced models, which is particularly important when the aim of inductive concept learning is to gain insight into the physical process that generated the data, rather than to predict the class of future cases. This is often discussed in the decision tree induction literature, where it is (crudely) measured by the number of nodes in the induced decision tree (Mingers, 1989).

### 9.4 HCI Cost of Incorporating Domain Knowledge

Several researchers have examined ways of embedding domain knowledge in a learning algorithm (Opitz and Shavlik, 1997). It has often been observed, in the context of expert system construction, that acquiring domain knowledge from a domain expert is a major bottleneck. We suppose that it would also be a bottleneck in the context of inductive concept learning, but we are not aware of any attempt to measure the cost.

## 10. Cost of Instability

When an induced model is used to gain understanding of the underlying process that generated the data, it is important that the model should be stable (Turney, 1995c; Domingos, 1998). By stability, we mean that, if two batches of data are generated from the same physical process, then the two corresponding induced models should be similar. If the two models are dissimilar, the learning algorithm is unstable. This is related to the scientific principle that experiments should be repeatable. Stability may be seen as a benefit and instability as a cost.

Stability may be increased by acquiring more data (using a larger training set) or by increasing the bias of the learning algorithm (Turney, 1995c). Acquiring more data can be costly (Section 8). Increasing the bias of an algorithm can increase the misclassification error rate (Section 2), unless the bias is suitable for the given learning task. Domingos (1998) presents a meta-learning algorithm, CMM, that can be used to trade off accuracy (Section 2), comprehensibility (Section 9.3), and stability.

## 11. Conclusion

In this paper, we have presented a taxonomy of types of cost in inductive concept learning. It is our hope that this taxonomy may serve to organize the literature on cost-sensitive learning and to inspire research into under-investigated types of cost.

We do not claim that this taxonomy is complete or unique. No doubt we have omitted important types of cost, and certainly other researches would prefer other taxonomies.

As we worked on this paper, it gradually became clear to us that the cost of misclassification errors occupies a unique position in the taxonomy. All of the other costs that we have discussed here can only be rationally evaluated in the context of the misclassification error cost (for the cost of intervention, the unwanted achievement cost is analogous to the misclassification error cost).

In decision theory (Pearl, 1988) and in the uncertainty in artificial intelligence literature (Pipitone *et al.*, 1991), test costs are generally considered in conjunction with misclassification error costs. However, in the inductive concept learning literature, it is striking that this has largely been overlooked. For example, before Turney (1995a), all of the papers on inductive concept learning with test costs did not consider test costs in the context of misclassification error costs (Núñez, 1988, 1991; Tan, 1991a, 1991b, 1993). Yet, if all test costs are greater than the misclassification error cost, then it is never rational to do any tests; and if the misclassification error cost is much greater than the cost of any test, then it is rational to do all of the tests, unless you are certain that they are irrelevant.

Similarly, as far as we know, none of the papers on active learning (Cohn *et al.*, 1995, 1996; Krogh and Vedelsby, 1995, Hasenjager and Ritter, 1998) consider the misclassification error cost, although we must know something about the misclassification error cost in order to rationally determine whether to pay the cost of the teacher.

## Acknowledgements

Thanks to Eibe Frank, Tom Fawcett, and Foster Provost for helpful comments on an earlier version of this paper.